%% file: PaperForReview.tex
\documentclass[10pt,twocolumn,letterpaper]{article}

\usepackage{cvpr}              %

\usepackage{graphicx}
\usepackage{amsmath}
\usepackage{amssymb}
\usepackage{booktabs}
\usepackage{mydefs}
\usepackage{lato}
\usepackage[table]{xcolor}
\usepackage[accsupp]{axessibility}

\usepackage[pagebackref,breaklinks,colorlinks]{hyperref}

\usepackage[capitalize]{cleveref}
\crefname{section}{Sec.}{Secs.}
\Crefname{section}{Section}{Sections}
\Crefname{table}{Table}{Tables}
\crefname{table}{Tab.}{Tabs.}

\def\cvprPaperID{14} %
\def\confName{CVPR}
\def\confYear{2023}

\begin{document}

\title{Visualizing Skiers' Trajectories in Monocular Videos}
\author{Matteo Dunnhofer \and
Luca Sordi \and
Christian Micheloni \and
Machine Learning and Perception Lab, University of Udine, Udine, Italy
}

\twocolumn[{%
\renewcommand\twocolumn[1][]{#1}%
\maketitle
\vspace{-0.9cm}
\begin{center}
    \centering
    \captionsetup{type=figure}
    \includegraphics[width=.95\linewidth]{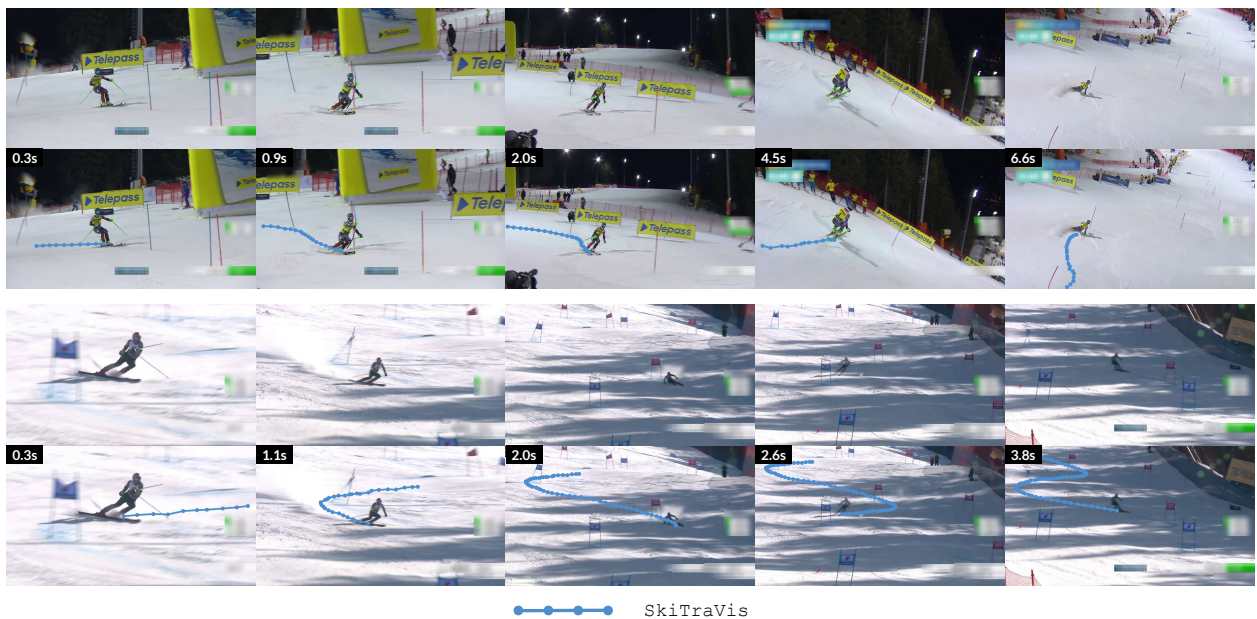}
    \captionof{figure}{\textbf{Visualization of skiers' trajectories in monocular videos.} We present \algoname, an algorithm to compute and visualize the \textcolor[HTML]{4B8ED1}{trajectory} performed by an alpine skier in videos acquired by a single monocular camera. \algoname\ takes in input a streaming video capturing a skier and visualizes its past trajectory in each frame. In this picture, the first and third rows show the input frames, while the second and fourth the trajectory computed by \algoname. The interval in seconds from the start of the video is also reported.}
    \label{fig:qualex}
\end{center}%
}]

\begin{abstract}
   \input{abstract.tex}
\end{abstract}

\input{introduction.tex}

\input{relatedwork.tex}

\input{methodology.tex}

\input{experimental.tex}

\input{results.tex}

\input{conclusions.tex}

\paragraph{\footnotesize Acknowledgments.} 
{\footnotesize
Research supported by the project between the University of Udine and the organizing committee of EYOF 2023 Friuli-Venezia Giulia.}

{\small
\bibliographystyle{ieee_fullname}
\bibliography{egbib}
}

\clearpage
\input{supp.tex}

\end{document}

%% file: abstract.tex
Trajectories are fundamental to winning in alpine skiing. Tools enabling the analysis of such curves can enhance the training activity and enrich broadcasting content. In this paper, we propose \algoname, an algorithm to visualize the sequence of points traversed by a skier during its performance. \algoname\ works on monocular videos and constitutes a pipeline of a visual tracker to model the skier's motion and of a frame correspondence module to estimate the camera's motion. The separation of the two motions enables the visualization of the trajectory according to the moving camera's perspective. We performed experiments on videos of real-world professional competitions to quantify the visualization error, the computational efficiency, as well as the applicability. Overall, the results achieved demonstrate the potential of our solution for broadcasting media enhancement and coach assistance.

%% file: introduction.tex
\section{Introduction}
Due to the popularity of alpine skiing \cite{SMTreport}, the competitive form of such a sport discipline is one of the most important markets of the winter sports industry \cite{nielsenreport}.
The international regulations of alpine skiing competitions \cite{icrfis} require the definition of a path of so-called turning gates down a hill. 
Differing in the overall length of the course and in the distance between the gates, four popular alpine skiing sub-disciplines are present nowadays: Slalom (SL), Giant Slalom (GS), Super-G (SG), and Downhill (DH). In each of these, the winning goal is the same: ski down through the gates of the designed course in the shortest time possible. 

To do so, an athlete should optimize its skiing trajectories along the track. Indeed, in alpine skiing the turns performed by an athlete form a trajectory of points on the slope surface that if optimized to reduce the distance with the turning gates can lead to saved time during the descent \cite{trajski,cai2020trajectory}. 
Toward such an objective, it becomes important to measure and analyze a trajectory in order to determine the specific points of the performance affecting the overall time taken. 
Machines able to perceive such evidence are valuable tools to obtain 
increased knowledge of the skiing performance.
The currently available solutions to measure a trajectory of an alpine skier make use of wearable devices (e.g. GNSS trackers, IMUs) put on the body or skis, and of 3D terrain models to position the data coming from the sensors at different time steps \cite{gilgien2013determination,kruger2010application}. 
The drawback of such an approach is the repeated time-consuming installation and calibration of the sensor device on the athlete's body, and the acquisition of a precise 
terrain model. 
Moreover, the contextual information influencing the skiing performance (e.g. positions of the turning gates, conditions of slope surface, weather, and visibility) is completely lost by such kinds of system. 
Research on training process improvement for skiing pointed out the opportunity for a post-performance video review to overcome such limitations \cite{supej2020methodological,ostrek2019existing}.

The importance of video-based performance analysis paves the way for training-assistive tools based on computer vision. Indeed, computer vision applied to videos capturing the athlete's performance is a valid option to measure skiing performance without the need for GNSS sensors or ground surface models \cite{sporri2016reasearch,ostrek2019existing}.
Previous work on this topic focused on 3D skier pose estimation \cite{SkiPose} on a motion capture system comprising six manually-operated mutually-calibrated pan-tilt-zoom cameras capturing a Giant Slalom track section of three gates. By means of multi-view human pose estimation algorithms \cite{rhodin2018learning,wandt2021canonpose,wandt2022elepose}, it was shown that the 3D pose of athletes can be reconstructed at different time steps, and from that time series, the skiing trajectory could be computed inside a virtual 3D environment.
Despite proving the suitability of computer vision for trajectory analysis, such work has limited applicability in practice because of the complex multi-camera setup.

In this paper, we aim for a different goal: to visualize, directly inside the frames of a capturing monocular video, the trajectory of a skier.
We describe the attempt to design and implement a computer vision algorithm to compute and visualize the trajectory in real competition conditions.
Our approach, we refer to as \algoname, enables the analysis of a trajectory 
in a qualitative way, meaning that the trajectory can be visualized in relation to the visual appearance of the context appearing in the video.
\algoname\ is inputted with just a single video with sufficient contextual details and works in a streaming fashion, i.e. in each video frame it outputs the perspective-correct trajectory exploiting the pixel information of the frame and the one occurring in the previous frames. 
Our solution decouples the skier's and the camera's motions between consecutive frames through a visual object tracker \cite{Stark} and an image key-point detection \cite{shitomasi} and tracking algorithm \cite{lk}.
The corresponding key-points are used by RANSAC \cite{ransac} to estimate the perspective transformation between the frames,  %
which in turn is used to map the previous motion of the skier to the  camera's changed perspective.
We evaluated \algoname\ on multiple clips taken from broadcasting videos capturing skier performance in real-world conditions. The qualitative and quantitative results achieved suggest the applicability of the method for augmented performance review and analysis during training sessions, and also for enriched broadcasting content. \algoname\ works not only for alpine skiing but even for snowboarding and for flight trajectory visualization in ski jumping and freestyle skiing.

%% file: relatedwork.tex
\section{Related Work}

\subsection{Computer Vision Applications in Skiing}
Thanks to the advancements in computer vision methodologies \cite{alexnet,resnet,fasterrcnn,openpose}, recently several applications have been possible for vision-based performance analysis in skiing.
In \cite{Skimovie}, a dataset was proposed to evaluate object detection and tracking algorithms to localize recreational alpine skiers in images and videos. 
The work described in \cite{Zhu2022} evaluated object detection and human pose estimation algorithms to recognize falls of alpine skiers. On a similar topic, it has been discussed about the combination of 2D pose detectors with kinematics models to distinguish normal skiing situations from out-of-balance and fall ones in pictures capturing competitive alpine skiers \cite{Zwolfer2021}.
\v{S}tepec et al. designed an algorithm based on 2D human pose trajectories to score the style of jumps in ski jumping from a monocular video \cite{Stepec_2022_WACV}. Concerning the same skiing discipline, Ludwig et al. designed improved vision transformer architectures \cite{tokenpose} to detect arbitrary key points on the human body \cite{Ludwig_2022_WACV} and on the skis \cite{Ludwig_2023_WACV} in still frames of the jumps.
For freestyle skiing and snowboarding, Wang et al. developed a systematic approach composed of a visual tracker, a human pose estimator, and a pose classifier, to evaluate the quality of aerial jumps in monocular videos \cite{wang2019ai}. In \cite{Matsumura2021}, research focused on the design of an algorithm to synchronize, spatially and temporally, two videos capturing snowboarders to compare the timing and spatial extent of their maneuvers.

The proposed \algoname\ shares some similarities with \cite{Stepec_2022_WACV,wang2019ai} in the systematic design based on a visual object tracker \cite{Stark,Dunnhofer2019,dunnhofer2021weakly,dunnhofer2022cocolot} to model the motion of a skier. However, differently from all the other works, here we tackle a different task that,
to the best of our knowledge, no other solution addressed before.

\subsection{Vision-based Trajectory Analysis in Sports}
Computer vision-based techniques have been successfully used in different sports disciplines to reconstruct the trajectory of various elements \cite{Chen2011,Maksai_2016_CVPR,kotera2019intra,Calandre2021,chen2018player}.
Chen et al. developed an algorithm to compute the 3D trajectory of a volleyball ball in a static-camera video by transforming the output of the ball's visual tracker with prior knowledge of the volleyball court \cite{Chen2011}. 
In similar capturing settings, Calandre et al. reconstructed the trajectory of table-tennis balls with a visual detection and tracking module, a registration  between image and table coordinates, and a kinematic ball motion model \cite{Calandre2021}.
In \cite{kotera2019intra}, the tracking-by-deblatting paradigm was introduced for better localize of -- hence compute better trajectories -- fast-moving objects such as table-tennis or tennis balls in static camera videos. 
In \cite{Maksai_2016_CVPR}, Mixed Integer Programming was used to track, under motion blur conditions, the position of volleyball, basketball, and soccer balls, with the goal of computing their motion trajectories.

In basketball analytics, different solutions have been proposed to reconstruct the trajectories of moving players \cite{Perse2009,Hu2011,Wei2013,chen2018player}. All these research endeavors presented a pipeline composed of a visual tracking algorithm to get player-specific localizations across video frames, and the application of a homography transformation between frame and court coordinates. Their differences lie in the tracking principles used, such as  histogram-based condensation \cite{Perse2009}, Cam-Shift \cite{Hu2011}, deformable part models and kalman filters \cite{Wei2013}, or deep learning methods \cite{chen2018player}, and in the homography estimation, based either on manual annotation \cite{Perse2009} or on the matching of court lines and boundaries \cite{Hu2011,Wei2013,chen2018player}.
Similar approaches have been also exploited in computer vision-based applications for soccer \cite{Honda_2022_CVPR} and ice-hockey \cite{VATS2023}.

The methodology described in this paper is similar to \cite{Perse2009,Hu2011,Wei2013,chen2018player} in the design of a system to track an athlete (in our case a skier), and to compute key-point correspondence from which to estimate a homography.
Differently from such solutions, we use a more recent transformer-based approach \cite{Stark} to implement the athlete tracking module, and we do not compute key-point matches based on some form of prior knowledge of the playing field. This is because alpine skiing is performed on a course that not only varies across competition locations but even across the different sections of the same mountain slope (e.g. in the width, steepness, and snow appearance). Hence, it is difficult to define a  model of the playing field as in soccer, basketball, or ice-hockey. For this motivation, in this work, we explore the usage of generic key-point correspondence algorithms \cite{SuperGlue,loftr,shitomasi,lk} to compute the homography that enables the computation of the skier trajectory.

%% file: methodology.tex
\begin{figure}[t]
\centering
  \includegraphics[width=.7\columnwidth]{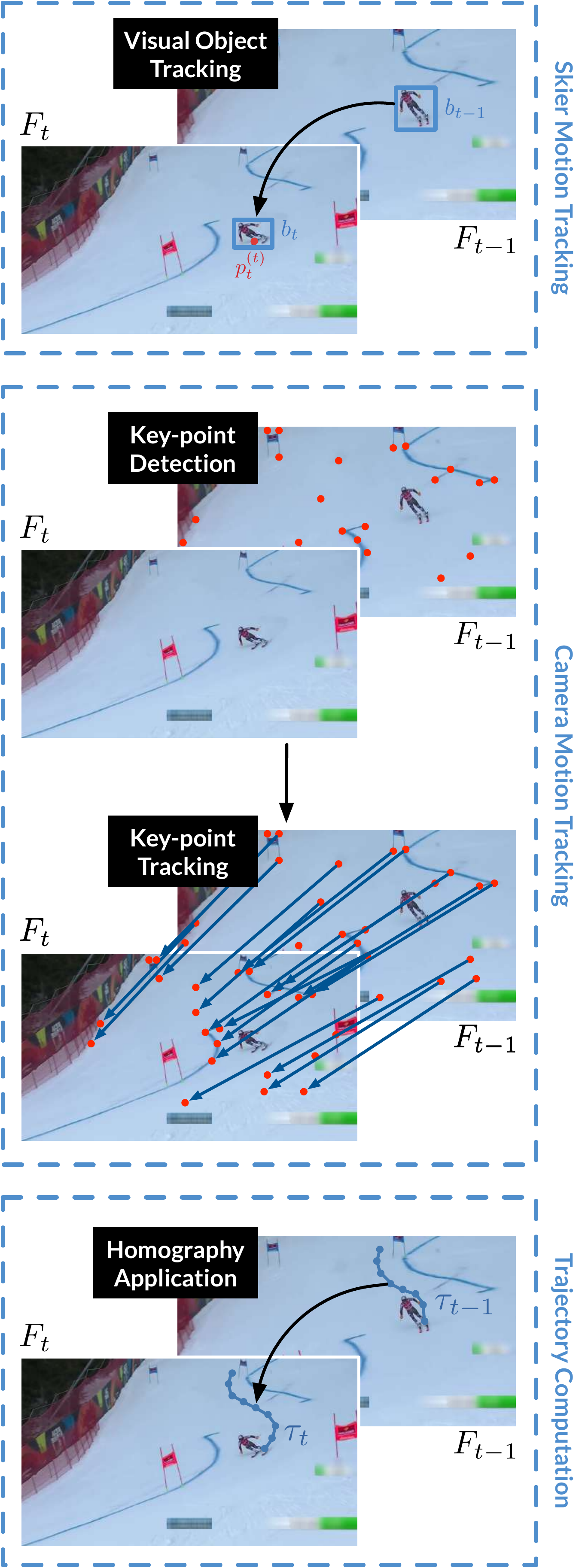}
  \caption{\textbf{Pipeline of }\algoname\textbf{.} 
  The algorithm processes each frame $\frame_t$ in an online fashion. A visual tracker is run to model the skier's motion between the previous frame $\frame_{t-1}$ and $\frame_t$. Detection and tracking of \textcolor[HTML]{DE3A3A}{static key-points} in the frames are run to estimate the camera's motion. Then, the homography between the two frames is computed and used to transform the previous trajectory \textcolor[HTML]{4B8ED1}{$\traj_{t-1}$} into \textcolor[HTML]{4B8ED1}{$\traj_{t}$} according to the perspective change.  
  } 
  \label{fig:pipeline}
\end{figure}

\section{Methodology}
\label{sec:method}
The goal of this paper is to develop an algorithm for augmenting a video capturing the performance of an alpine skier with the visualization of the trajectory he/she skied from his/her position in the first frame up to his/her position in each other frame. 
Qualitative examples of our intention are presented in Figure \ref{fig:qualex}.

\subsection{Preliminaries}
The videos given as input to our \algoname\ are considered to capture the performance of an individual skier while he/she is constantly visible (at least partially) in the scene. 
We do not put any constraints on the configuration (intrinsic and extrinsic parameters) of the camera that captured the videos, if not the presence of a sufficient amount of visual features in the scene (more details about this will be given in Section \ref{sec:experimental}).
Formally, we consider a video
$\video = \big\{ \frame_t \in \images \big\}_{t=0}^{T}$
as a sequence of $T \in \mathbb{N}$ frames $\frame_t$, where $\images =  \{0,\cdots,255\}^{w \times h \times 3}$ is the space of RGB images.
We use $\point^{(t)}_t = (x^{(t)}_t, y^{(t)}_t)$ to denote the coordinates of the point that summarizes the position of the athlete in the image coordinate system of $\frame_t$ (e.g. the point of contact between the skier and the snow surface).

\paragraph{Problem Formulation.} 
The goal is to compute, at each frame $\frame_t$, a trajectory 
\begin{align}
\traj_t = \big\{ \point^{(t)}_i \big\}_{i=0}^{t},
\end{align}
which is the sequence of points $\point^{(t)}_i$ traversed by the athlete in the 2D space of previous frames, from $\frame_0$ up to $\frame_t$. 
To respect an online processing modality, the algorithm is not allowed to process $\frame_{\widehat{t}}, \widehat{t} > t$ 
but just the pixel information available in $\frame_t$ and eventual computations performed in preceding frames.
After its computation, the points in $\traj_t$ are used to highlight the pixels of $\frame_t$, hence showing the trajectory performed by the skier. The online setting aligns with the real-time requirements of different applications (e.g. broadcasting and coach assistance) because it does not require waiting for the athlete's exercise to be terminated for the trajectory to be produced.

\subsection{Pipeline of \algoname}
Figure \ref{fig:pipeline} depicts the pipeline of the proposed \algoname. 
The algorithm processes each $\frame_t$ online according to their temporal order. $\frame_t$ is first given to a visual object tracking algorithm designed to model the motion of the skier and to provide its position $\point^{(t)}_t$. Then, the pipeline 
estimates the camera's motion between the consecutive frames $\frame_{t-1}, \frame_t$ by computing the homography transformation $\homo \in \mathbb{R}^{3\times3}$. This is found by detecting and tracking static key-points present in the two frames and by using RANSAC on the corresponding key-points. 
$\homo$ is used to map the points of the trajectory available in the previous frame, $\tau_{t-1}$, into the coordinate system of $\frame_t$. After that, $\point^{(t)}_t$ is appended to such a transformed trajectory to obtain $\tau_t$, the list of all the points traversed by the athlete with respect to $\frame_t$'s perspective. 
We now describe the different components of \algoname\ in more detail.

\begin{figure}[t]
\centering
  \includegraphics[width=\linewidth]{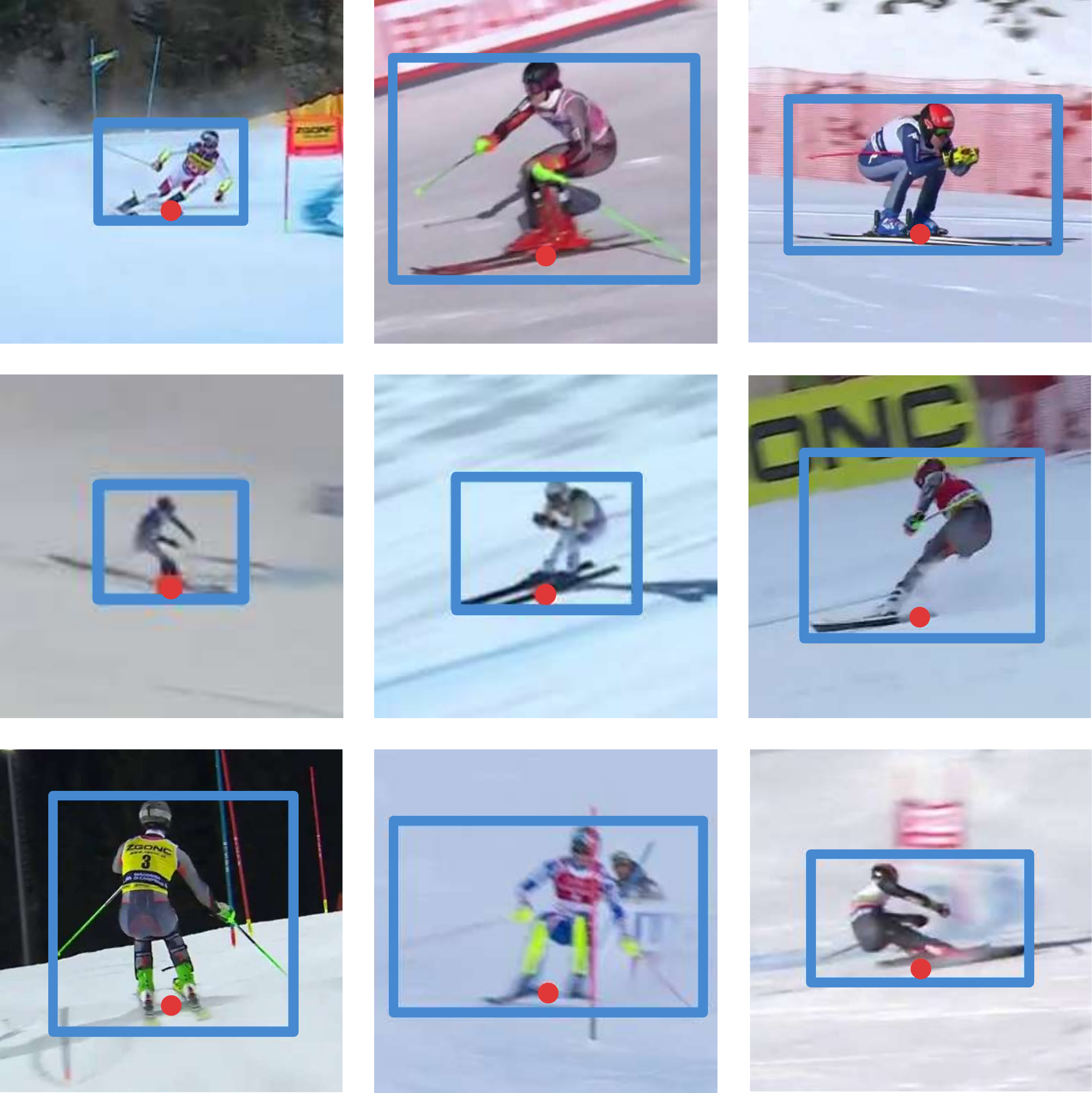}
  \caption{\textbf{The points forming a trajectory.} The points \textcolor[HTML]{DE3A3A}{$\point^{(t)}_t$} that compose the trajectory $\tau_t$ (represented here as \textcolor[HTML]{DE3A3A}{red dots}) are extracted from the bounding-box \textcolor[HTML]{4B8ED1}{$\bbox_t$} (represented by the \textcolor[HTML]{4B8ED1}{blue rectangles}) outputted by the skier tracking algorithm. The points roughly denote the skiers' feet' position.} 
  \label{fig:feetpt}
\end{figure}

\paragraph{Skier Motion Tracking.}
The first step of the pipeline is to exploit a visual object tracker 
\cite{VOT2022,Dunnhofer2020accv,dunnhofer2022cocolot} 
to track the motion of the athlete across all the frames up to $\frame_t$. For this, we used a tracker outputting a bounding-box $\bbox_t = (x^{(\bbox)}_t,y^{(\bbox)}_t,w^{(\bbox)}_t,h^{(\bbox)}_t) \in \mathbb{R}^4$ enclosing the athlete's appearance at every $\frame_t$. $x^{(\bbox)}_t,y^{(\bbox)}_t$ represent the coordinates of the top-left corner of the box while $w^{(\bbox)}_t,h^{(\bbox)}_t$ estimate the width and height.
We consider the position of the skier $\point^{(t)}_t = (x^{(t)}_t, y^{(t)}_t)$ in $\frame_t$ as 
\begin{align}
    x^{(t)}_t &= x^{(\bbox)}_t + w^{(\bbox)}_t \cdot 0.5\\
    y^{(t)}_t &= y^{(\bbox)}_t + h^{(\bbox)}_t \cdot k\\
    k &\in [0,1].
\end{align}
 Given an accurate bounding-box and $k = 0.9$, we assume $\point^{(t)}_t$ to approximate the closest point of contact between the skier's feet and the snow, as shown by Figure \ref{fig:feetpt}.\footnote{More information is given in Appendix \ref{sec:points} of the supplementary document.}
The tracker is initialized in the first frame $\frame_0$ of the video with the bounding-box $\bbox_0$ that outlines the appearance of the target skier. Such a piece of information can be obtained by asking a human operator to provide the bounding-box for the athlete of interest via some user-friendly annotation system or by a skier-specific detection algorithm. %
Based on experiments, we found the state-of-the-art transformer-based single object tracker STARK \cite{Stark} to provide a great balance between accuracy and efficiency. We fine-tuned the pre-trained publicly-available model executing the original training methodology on video data specifically designed for alpine skier tracking.

\paragraph{Camera Motion Tracking.}
To render the trajectory of the skier $\traj_{t-1}$ into $\frame_t$, we need to compute the perspective transformation occurring between $\frame_{t-1}$ and $\frame_t$.
To implement this step we make use of the homography matrix $\homo$ mapping the pixel coordinates of $\frame_{t-1}$ into $\frame_t$.
To compute $\homo$, we match static key-points in the fields of view of $\frame_{t-1}$ and $\frame_t$ through an off-the-shelf generic key-point detector and tracker. Key-points are first detected in $\frame_{t-1}$, and then the correspondences of such points are found in $\frame_t$ by local tracking. This procedure computes an alignment between the same semantic points of the images that express how static objects in the scene have apparently moved between the frames. Since the objects are static and do not actually move, the camera's motion is quantified \cite{multiviewgeometry,stitching}. 
Based on experiments, we have found traditional methods \cite{shitomasi,lk} to work the best. The method of Shi-Tomasi \cite{shitomasi} was used to detect visual features in $\frame_{t-1}$, and the Lucas-Kanade's optical flow method \cite{lk} to track such visual descriptors in $\frame_t$.
The usage of such off-the-shelf instances is motivated by the fact that alpine skiing is performed on courses that change a lot, across competition locations and even across the different sections of the same mountain slope. Hence, it is not trivial to define a generally applicable model of the playing field, which we leave for future investigations.

\paragraph{Trajectory Computation.}
With the alignments of corresponding key-points, we compute the homography matrix $\homo$ through the RANSAC algorithm \cite{ransac}, which applies an iterative optimization procedure on the correspondences in order to find the best homography matrix that explains them.
We excluded the key-points in $\frame_t$ within the bounding-box $\bbox_t$ because they belong to the skier which is a non-static object. If present, we also discarded all the key-points lying on the superimposed virtual graphics showing the characteristics of the athlete's performance (e.g. the running time).
After  $\homo$ is computed, it is used to map the points $\point^{(t-1)}_i$ of $\traj_{t-1}$ in the new frame $\frame_t$. 
In more detail, at each $\frame_t$, $\traj_{t-1}$ consists of all the skier positions tracked in the preceding $t-1$ frames and transformed according to the perspective of $\frame_{t-1}$. 
The trajectory $\traj_t$ for the latest $\frame_t$ is obtained by the multiplication of the homogeneous coordinates of each $\point^{(t-1)}_i \in \traj_{t-1}$ with the homography matrix, such that  
\begin{align}
    \label{eq:traj}
    &\traj_t = \big\{ \point^{(t)}_i \big\}_{i=0}^{t}, \\
    \label{eq:extremes}
    &\point^{(t)}_i = [x^{(t)}_i, y^{(t)}_i] : 0 \leq x^{(t)}_i < w, 0 \leq y^{(t)}_i < h, \\
    &[x^{(t)}_i, y^{(t)}_i, 1] = [x^{(t-1)}_i,y^{(t-1)}_i, 1]^{\mathbf{T}} \cdot \homo, \\
    &[x^{(t-1)}_i, y^{(t-1)}_i] = \point^{(t-1)}_i. 
\end{align}

As can be noticed by Eq. (\ref{eq:traj}), $\traj_t$ also includes  the point $\point^{(t)}_t$ extracted by the tracker's bounding-box for $\frame_t$. This is achieved by appending such a point to $\traj_t$ after the application of $\homo$ to $\traj_{t-1}$. Eq. (\ref{eq:extremes}) specifies that only the transformed points falling inside the $\frame_t$'s image coordinate system are retained in  $\traj_t$.

More concretely, at the first time step in which the frame matching procedure is executed, i.e. $t = 1$, $\traj_{0} = \{ \point^{(0)}_0 \}$ consists only of the point extracted by the bounding-box (detected or manually labeled) highlighting the target skier in $\frame_0$. To obtain the transformation of $\point^{(0)}_0$ in $\frame_1$, its homogeneous representation is multiplied by $\mathbf{H}_1$ to compute the homogeneous vector $[x^{(1)}_0, y^{(1)}_0, 1]$, from which $\point^{(1)}_0$ is extracted. $\traj_1 = \{ \point^{(1)}_0, \point^{(1)}_1 \}$ is obtained by appending $\point^{(1)}_1$. The process is repeated until the end of the video.

%% file: experimental.tex
\section{Experimental Settings\protect\footnote{Detailed information is provided in the Appendix \ref{sec:dataappendix}.}}
\label{sec:experimental}

\subsection{Data and Annotations}
\label{sec:data}
To determine the quality of \algoname's trajectories, we make use of a dedicated evaluation dataset representing real-world application scenarios. The videos belonging to this dataset appear in the test-set of SkiTB, a video dataset designed for the development of athlete tracking methods in different skiing disciplines.\footnote{SkiTB is developed by our research team and it will be released soon with a separate publication.} 
Such a dataset features 100 multi-camera broadcasting videos (60 videos for training and 40 for testing) that capture completely, from the start to the finish line, the performance of professional alpine skiers. Each frame of each video is manually annotated with a bounding-box containing the appearance of the athlete's body and equipment. 
 Manually annotating the trajectories performed by skiers in each frame of the video clips is a challenging task other than very time-consuming. Indeed, it is hard to visually identify the specific points at which the skier passed since the whitish appearance of the snow's surface does not show easily identifiable reference points to localize them. %
 We hence exploited an automatic annotation procedure followed by manual verification. We executed the state-of-the-art LOFTR image matching algorithm \cite{loftr} to detect and match key-points of consecutive frames of SkiTB's monocular test videos. RANSAC \cite{ransac} was then used to compute homographies on the matches. The homographies were used to map the skier localization points extracted from the manual bounding-box annotations into trajectories for each frame, in a similar fashion as described for \algoname. The videos with the generated trajectories have been visually assessed and selected by our team, after being instructed by two professional alpine skiing coaches on meaningful trajectory visualizations. 
 Overall, this procedure allowed us to obtain 233 monocular clips whose characteristics are reported in Table \ref{tab:stats}.

\input{tables/stats.tex}

 \begin{figure}[t]
\centering
  \includegraphics[width=\linewidth]{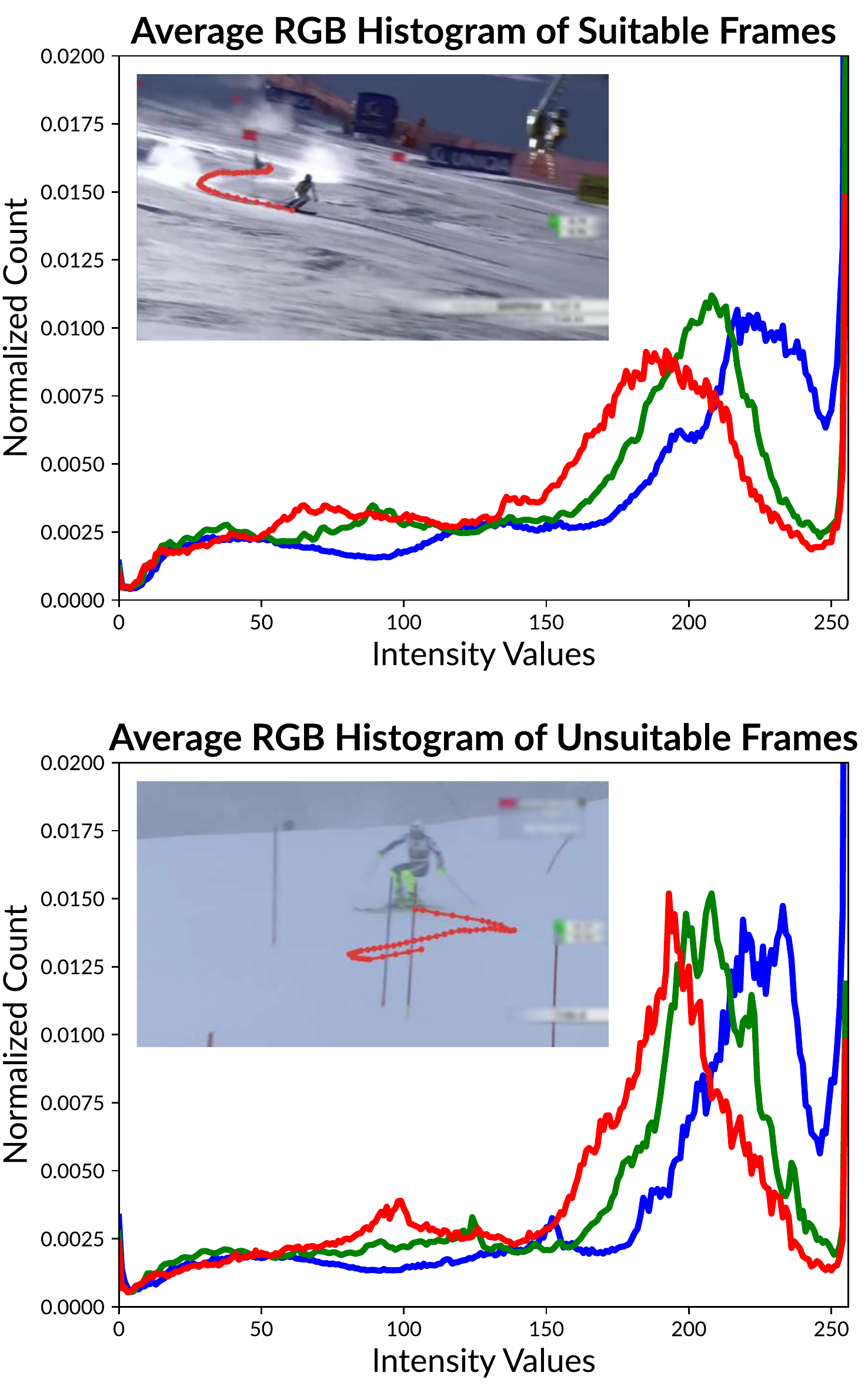}
  \caption{\textbf{RGB color distributions for the applicability of }\algoname\textbf{.} The top plot shows the average RGB values distribution and an example (with the \textcolor[HTML]{DE3A3A}{generated reference trajectory}) of the frames belonging to the evaluation set used for \algoname, while the bottom one shows the RGB values distributions and an example (with a \textcolor[HTML]{DE3A3A}{wrongly generated trajectory}) of unsuitable frames. As can be noticed, suitable clips include frames that have a larger image contrast.} 
  \label{fig:hist}
\end{figure}
 
 Leveraging the proposed semi-automatic annotation protocol limits the validity of the proposed \algoname\ to the videos whose frames are successfully processed by LOFTR \cite{loftr}. 
 We tried to quantify the conditions in which such an algorithm works by analyzing the difference between used and discarded video clips. As can be noticed by the width of the RGB histograms in Figure \ref{fig:hist}, the suitable videos include frames with a generally larger image contrast. We also evaluated the number of features per-frame discovered by the Shi-Tomasi algorithm \cite{shitomasi}, resulting in an average number of 1430 (median 1219) for suitable clips and 1126 (median 811) for unsuitable ones.
 These differences are mostly due to the appearance of winter scenarios in images, which provide whitish textures containing few visual anchors that feature detection algorithms can exploit.

\subsection{Evaluation Protocol}
The protocol used to execute \algoname\ for evaluation is similar to the One-Pass Evaluation for visual tracking \cite{OTB}. The algorithm is run on each video separately, and it is given the first frame and the respective ground-truth bounding-box to initialize the visual tracker on the skier. Then, \algoname\ is inputted with each of the remaining frames $\frame_t$ iteratively, it processes the frames according to the described methodology and returns each $\traj_t$. To compute quantitative results, each $\traj_t$ is compared with the respective reference trajectory $\trajgt_t$, 
created as described in the previous sub-section. In addition, we evaluate the bounding-boxes $\bbox_t$ predicted by the visual tracker and the homographies $\homo$ generated after the frame matching step, in order to understand the impact of different implementations.

\subsection{Evaluation Measures}

To quantify the distance between $\traj_t$ and $\trajgt_t$, we compute the average Euclidean pixel distance between the points of the reference $\trajgt_t$ with the respective, ordered based on the insertion time step, of $\traj_t$. We refer to this measure as Mean Per Point Trajectory Error (\mppte).
In addition, we report the pixel values of Dynamic Time Warping (\dtw) \cite{dtw}, a popular measure in trajectory analysis.
To evaluate the skier tracking, we employ the Area Under the Curve (\auc) of the success plot which is equivalent to the average bounding-box overlap \cite{OTB}.
For the frame matching module, we evaluate the mean squared error (\mse) between the estimated homographies and the respective LOFTR-validated ones.

Besides the error, we also rated the efficiency of \algoname\ under different configurations of visual trackers and frame matching methods. For both categories, we report the average module-specific running speed in milliseconds (\ms) and in Frames-Per-Second (\fps), and the average delay in seconds (\timediff) between the time instant of the video clip's end and the end instant of the whole \algoname\ pipeline's processing time. This value measures the average overall delay in visualizing the trajectory for the entire video clip, and it can be used to understand how much time has to be waited to obtain the trajectory after the observation of the skier's performance.

\subsection{Evaluated Implementations}
To the best of our knowledge, \algoname\ is the first algorithm for skier trajectory visualization in monocular videos, and there is no comparable solution present in the literature. Thus, to provide useful insights, we evaluate the performance of our algorithm under different configurations of state-of-the-art visual trackers and frame correspondence algorithms.
For the visual tracker, we evaluated: the deep learning-based methodologies STARK \cite{Stark}, SuperDiMP \cite{DiMP,VOT2020}, SiamRPN++ \cite{siamrpn}, trained for generic object tracking; and the traditional method MOSSE \cite{MOSSE}. We also fine-tuned STARK with the aforementioned SkiTB training data.
For key-point detection and matching, we evaluated: the learning-based method of SuperPoint (SP) \cite{SuperPoint} and SuperGlue (SG) \cite{SuperGlue}, and the Recurrent All-pairs Field Transform
 (RAFT) dense optical flow method \cite{raft};
and the traditional methods as the Shi-Tomasi (ST) \cite{shitomasi} detector with the optical flow algorithm Lucas-Kanade \cite{lk},
and the Oriented FAST and Rotated BRIEF (ORB) descriptor \cite{orb} with the brute force (BF) matcher.
Code was implemented in Python and run on a machine with an Intel Xeon E5-2690 v4 @ 2.60GHz CPU, 320 GB of RAM, and an NVIDIA TITAN V GPU.

%% file: tables/stats.tex
\begin{table}[t]

\fontsize{7}{8}\selectfont
	\centering
	\caption{\textbf{Statistics of the dataset used to evaluate } \algoname\textbf{.} The first block of rows gives details about the characteristics of the video clips, while the second provides information about the featured real-world application scenarios.}
	\label{tab:stats}
        \rowcolors{1}{tblrowcolor2}{tblrowcolor1}
	\begin{tabular}{l | c  }
		\toprule
        \# monocular video clips & 233 \\
		\# frames & 25783 \\
        Video clip length in \# frames (min, avg, max) & 30, 111, 354 \\
        Video clip length in seconds (min, avg, max) & 1.0, 3.7, 11.8 \\
        Resolution, frame-rate & 720p, 30 FPS \\
        $\traj_t$ length in \# points (min, avg, max) & 1, 25, 229 \\
        \midrule
        
        \# clips for alpine skiing sub-discipline (SL, GS, SG, DH) & 35, 55, 71, 72 \\
        \# athletes (men, women) & 23 (13, 10) \\
        \# locations & 29 \\
        \# weather conditions (clear, overcast, fog, snowing) & 200, 30, 2, 1 \\
		
		\bottomrule		
\end{tabular}

\end{table}

%% file: results.tex
\input{tables/tracking.tex}

\section{Discussion}
\begin{figure*}[t]
\centering
  \includegraphics[width=\linewidth]{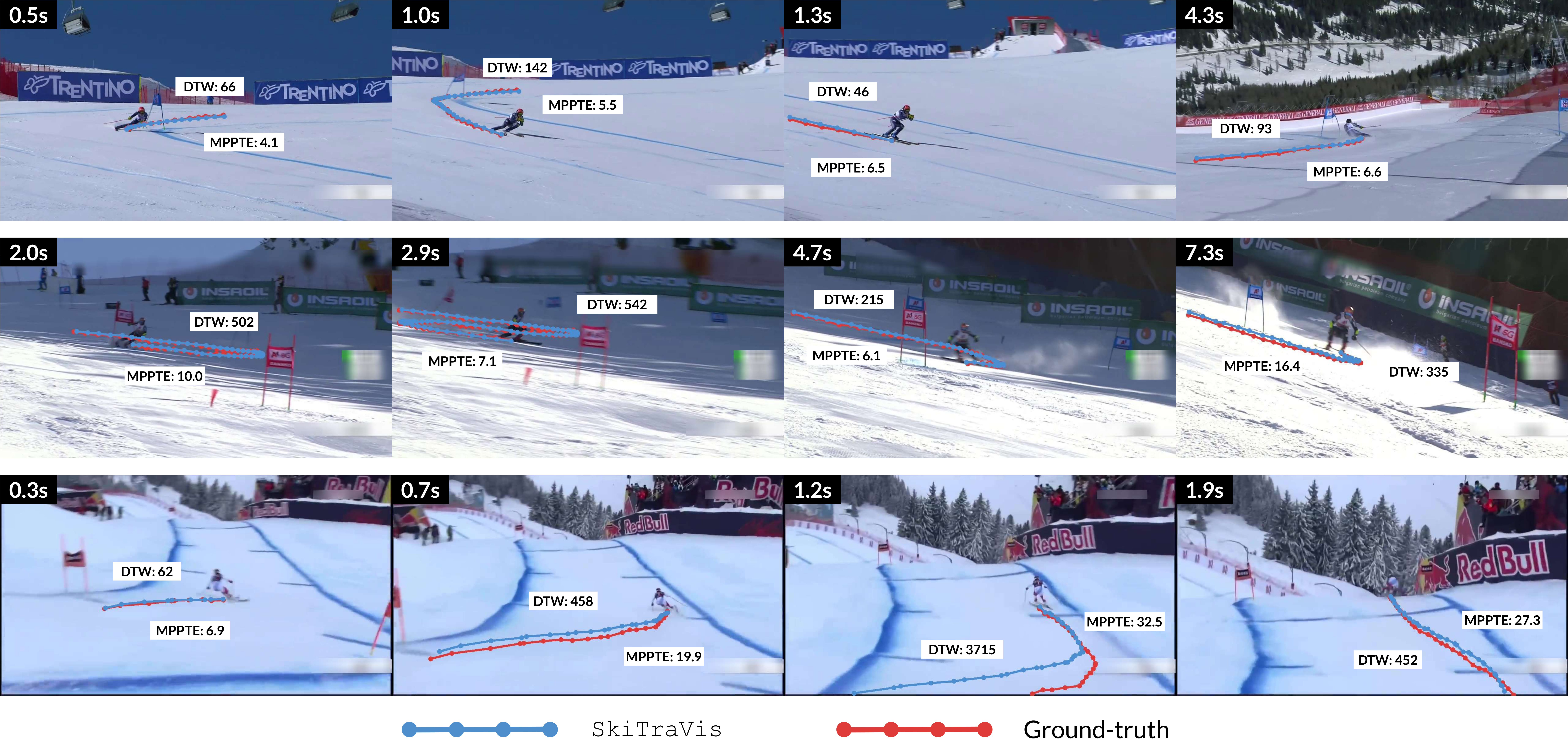}
  \caption{\textbf{Visualizations of trajectory errors.} Here we provide visual demonstrations of how the \mppte\ and \dtw\ values represent the quality of the \textcolor[HTML]{4B8ED1}{trajectory} visualizations. The first two rows show frames where the reconstruction is visually consistent, while the last row presents video frames with a large error. } 
  \label{fig:errorvis}
\end{figure*}

In this section, we discuss the results achieved by \algoname. Overall, by looking at the numbers in Table \ref{tab:tracking} and \ref{tab:matching}, we observe that the configuration with the fine-tuned STARK-ft tracker and the ST + LK key-point detector and tracker has the best balance between trajectory error and processing speed. A visualization of the \mppte\ and \dtw\ errors, as presented in Figure \ref{fig:errorvis} demonstrate that \mppte\ and \dtw\ values lower than 15 and 400 pixels lead to a consistent perception of the trajectory. 
Some demonstrative videos are available at \videolink. 
The \mppte\ score is less than the 10\% of the average skier appearance's height which is 123 pixels (computed from the manually annotated bounding-boxes).
The running time of the STARK-ft ST + LK configuration has an average \timediff\ of 6 seconds which is promising for usage in broadcasting applications (e.g. for replays) and during training activities (as the real-time requirement is not so strict).

\paragraph{Skier Motion Tracking.}
Table \ref{tab:tracking} reports the accuracy of \algoname\ when different visual trackers are employed. Of the models trained for generic object tracking (shown in rows 3-6), STARK results in the best performing under the \mppte\ and \auc\ measures, while SuperDiMP achieves the lowest \dtw\ score. SiamRPN++ and MOSSE fall behind in all the scores, apart from the processing speed where they result better. Considering this aspect, it is worth noticing that STARK presents a great balance between accuracy and efficiency. For this motivation, we fine-tuned it (as STARK-ft, second row of Table \ref{tab:tracking}) with the alpine skier-specific training-set of SkiTB. Such a version reduces the \mppte\ and \dtw\ of 37\% and 35\%, respectively, and presents an \mppte\ error higher only of 2.3 pixels with respect to the trajectories extracted by the manually labeled bounding-boxes (denoted as Oracle, first row).

\input{tables/matching.tex}

\paragraph{Camera Motion Tracking.} Table \ref{tab:matching} presents the performance of \algoname\ under different frame matching strategies. The ST detector and the LK optical-flow method (ST + LK, row two) results in the closest to LOFTR in the quality of the trajectories, and the best in the computational efficiency. The SG strategy (SP + SG, row three) follows shortly in \mppte\ and \dtw, while it is much slower in the computation. We observe a good trajectory computation by ST + LK even though its computed homographies present a larger \mse\ to those of the latter strategy. 
The remaining tested methods, ORB + BF and RAFT, show a much larger error in all the considered measures.
Table \ref{tab:matching} also reports the results achieved with the LOFTR method (row one) used to generate the reference trajectories. When run in this configuration with STARK-ft as the skier tracker, the \mppte\ and \dtw\ decrease by 60\% and 68\% respectively. By comparing these improvements with those given by the Oracle tracker, we can conclude that most of the \algoname's error is due to the frame matching module. 
Due to its good \mppte\ and \dtw\ values, as well as considering the discussion in Section \ref{sec:data}, the results of the STARK-ft and LOFTR configuration demonstrate its applicability in practice, but when the processing time is not an issue since LOFTR is the slowest frame matching strategy.

\begin{figure}[t]
\centering
  \includegraphics[width=\linewidth]{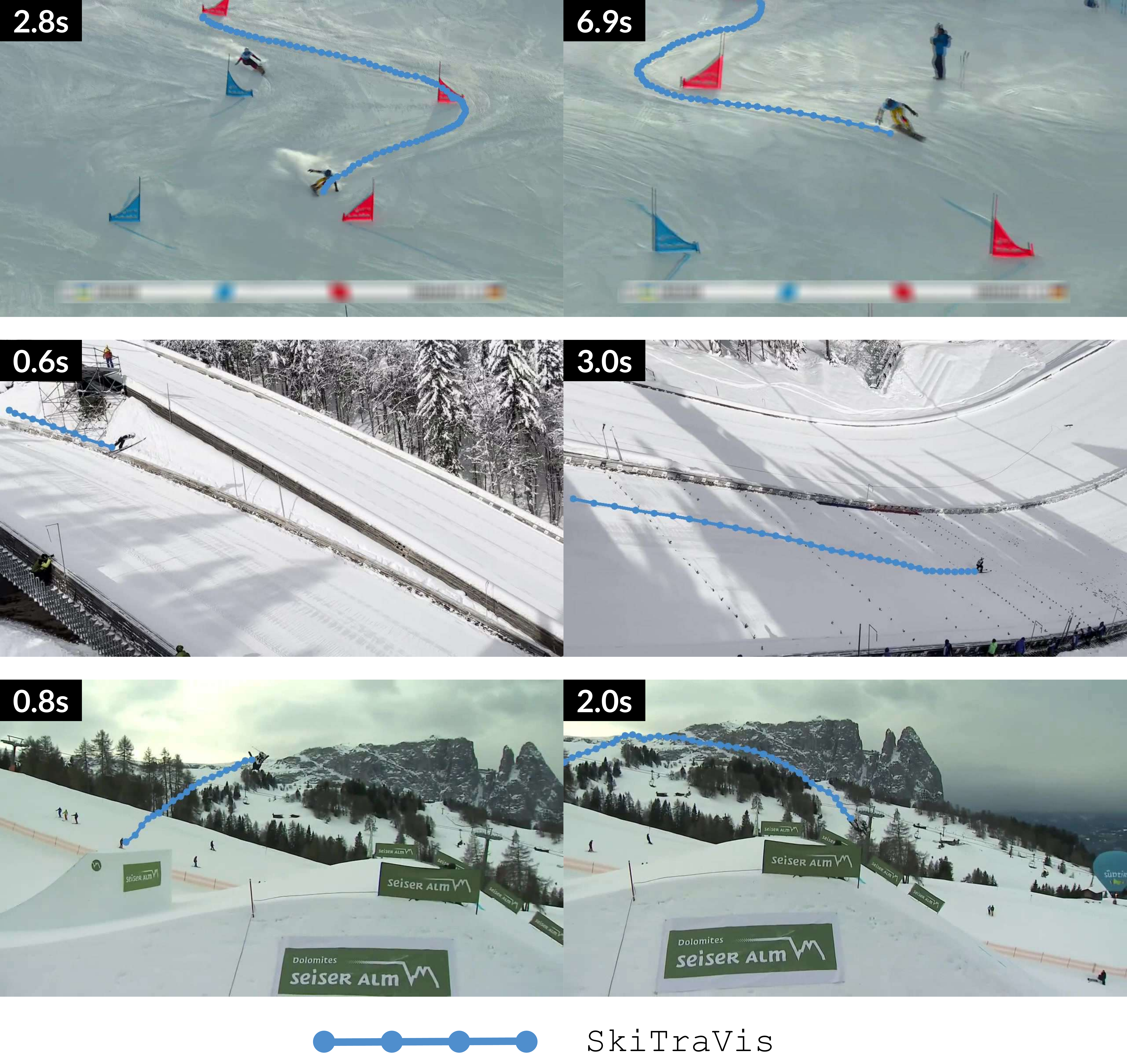}
  \caption{\textbf{Generalization to other skiing disciplines.} We found \algoname\ to work well for the generation of \textcolor[HTML]{4B8ED1}{trajectories} in other skiing disciplines such as snowboarding, and for the visualization of the \textcolor[HTML]{4B8ED1}{flight trajectory} in ski jumping and freestyle skiing. }
  \label{fig:generalization}
\end{figure}

\paragraph{Generalization.} 
Our algorithm generalizes to other skiing disciplines. Figure \ref{fig:generalization} shows qualitative examples of \algoname\ applied to the generation of the trajectory performed by alpine snowboarders, and to the visualization of the flight trajectory in ski jumping and freestyle skiing.

%% file: tables/tracking.tex
\begin{table}[t]

\fontsize{7}{8}\selectfont
	\centering
	\caption{\algoname\textbf{'s performance with different visual trackers for skier motion modeling.} The STARK tracker \cite{Stark} fine-tuned for skier tracking presents great accuracy coupled with a good running time. In all of these experiments, the ST + LK key-point detector and tracker were used.}
	\label{tab:tracking}
	\setlength\tabcolsep{.19cm}
    \rowcolors{2}{tblrowcolor2}{tblrowcolor1}
	\begin{tabular}{l | c c | c  | c c c }
		\toprule
		Tracker & \mppte & \dtw & \auc & \ms & \fps & \timediff \\
        \midrule
        Oracle & 9.8 & 278 & 100 & - & - & - \\
        STARK-ft & 12.1 & 313 & 90.1 & 29 & 34 & 6 \\
        STARK & 19.3 & 484 & 73.2  & 29 & 34 & 6 \\
        SuperDiMP & 21.2 & 479 & 67.6 & 89 & 12 & 11 \\
        SiamRPN++ & 31.6 & 791 & 62.8 & 20 & 49 & 5 \\
        MOSSE & 76.1 & 1075 & 37.0 & 9 & 117 & 3 \\

		\bottomrule		
\end{tabular}

\end{table}

%% file: tables/matching.tex
\begin{table}[t]

\fontsize{7}{8}\selectfont
	\centering
	\caption{\algoname\textbf{'s performance with different frame matching algorithms.} The ST + LK \cite{shitomasi,lk} combination shows the greatest balance between accuracy and processing speed. In all of these experiments, the STARK-ft visual tracker was used.}
	\label{tab:matching}
	\setlength\tabcolsep{.11cm}
    \rowcolors{2}{tblrowcolor2}{tblrowcolor1}
	\begin{tabular}{l | c c | c | c c c }
		\toprule
		Detection \& Matching & \mppte & \dtw & \mse & \ms & \fps & \timediff \\
        \midrule
        LOFTR & 4.8 & 101 & 0 & 801 & 1 & 90 \\
        ST + LK & 12.1 & 313 & 9952 & 42 & 24 & 6 \\
        SP + SG & 15.1 & 352 & 25.7 & 135 & 7 & 15 \\
        ORB + BF & 28.7 & 836 & 111732 & 783 & 1 & 84 \\
        RAFT & 39.8 & 807 & 22 & 641 & 2 & 73 \\
	
		\bottomrule		
\end{tabular}

\end{table}

%% file: conclusions.tex
\section{Conclusions}
In this paper, we present \algoname, an algorithm to visualize the trajectories performed on an alpine skier in a monocular video. The solution uses a skier-specific visual tracking algorithm to model its motion, and an image-generic key-point-based frame matching algorithm to estimate the camera motion and the consequent transformation between the video frames. Thanks to the separation of the two motions, it is possible to relocate the previous skier positions according to the captured perspective, hence obtaining a visualization of the trajectory performed by the athlete. The experimental campaign focused on analyzing the algorithm's accuracy and efficiency, as well as the conditions of applicability. Under some circumstances, we believe that \algoname\ can be used in broadcasting and coach assistive applications, and more generally, it will serve as a baseline for future research on trajectory visualization for computer vision-based skiing analytics.

%% file: supp.tex
\twocolumn[{%
\centering
\vspace{1em}
{\Large \textbf{Visualizing Skiers' Trajectories in Monocular Videos}} \\
\vspace{.5em}
{\large  Supplementary Document} \\
\vspace{1em}
Matteo Dunnhofer, Luca Sordi, Christian Micheloni \\
Corresponding author e-mail: \href{mailto:matteo.dunnhofer@uniud.it}{\texttt{matteo.dunnhofer@uniud.it}}
\vspace{2.5em}
}]

\appendix
\section{Further Motivations and Details about \algoname}

\subsection{Trajectory Points $\point^{(t)}_t$}
\label{sec:points}
To represent the skier localization point $\point^{(t)}_t$, we did not use a human skeleton parsing algorithm, such as a 2D human pose detector \cite{openpose,alphapose}, to not increase the complexity and the efficiency of the \algoname\ pipeline. In fact, such a procedure would have required the execution of an additional deep neural network at the cost of incrementing the overall time to visualize the trajectory. Moreover, even though alpine skier-specific pose detectors have been studied \cite{SkiPose}, we found them committing displacement errors in the prediction of the human and equipment key-points close to the ground surface (e.g. for prediction feet' or skis' positions). For instance, AlphaPose \cite{alphapose} fine-tuned and tested, respectively, on the SkiPose2D training and test datasets \cite{SkiPose} achieves a Mean Per Joint Prediction Error (MPJPE) of 13.2 pixels at the skier's feet and of 9.4 pixels in the upper body parts. By calculating the average pixel distance between $\point^{(t)}_t$ and its corresponding reference point (i.e. the point computed following Eq. (2,3,4) on the manually annotated bounding-box) we obtain a value of 5.7 pixels, which is much lower than the errors committed by the skier pose detector. These results demonstrate that the bounding-box tracker is more robust in localizing the point $\point^{(t)}_t$ in practice.
Nevertheless, we believe that \algoname\ does not require substantial changes in the pipeline to compute more semantically meaningful $\point^{(t)}_t$, since 2D human pose detector could be just run on the image patches extracted from the bounding-boxes predicted by the visual tracker.
 We hence leave the integration of 2D pose detectors for future work. Moreover, the selection of values $k < 0.9$ (e.g. $k = 0.5$) could be used to move the trajectory toward representing the athlete's center of mass, which is of interest for biomechanical analysis \cite{supej2020methodological}, but it should be noted that in 2D images such a point does not fall at the same depth level in which the homography is computed. Indeed, in nearby locations to the center of mass, the homography is computed by exploiting static key-points in the background that lie on the snow's surface behind the skier. Hence, transforming the eventual $\point^{(t)}_t$ with such a homography would shift the point by a pixel amount that does not respect the camera motion at the depth level at which the skier is actually localized.

\section{Experimental Details}
\label{sec:dataappendix}

\begin{figure*}[t]
\centering
  \includegraphics[width=\linewidth]{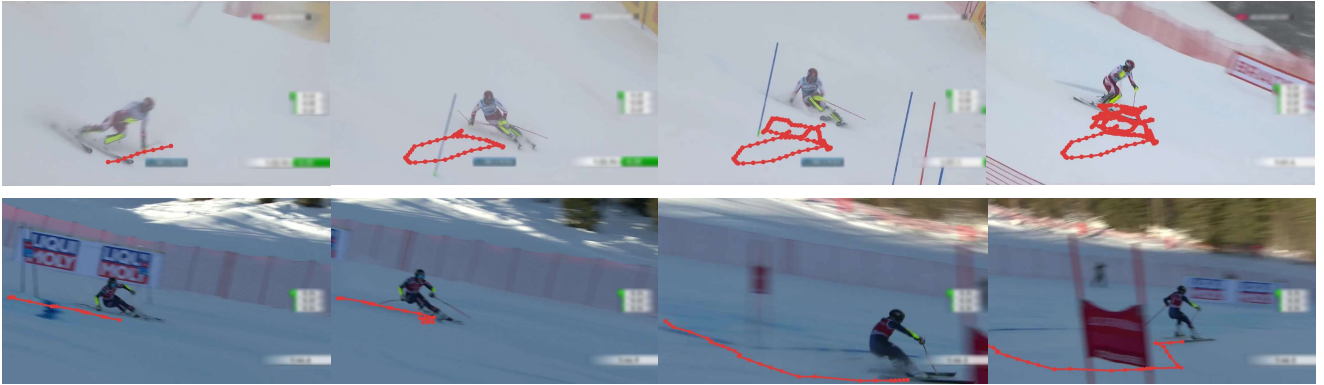}
  \caption{\textbf{Suitable and unsuitable clips for trajectory visualization.} This figure extends the plots of Figure \ref{fig:hist} of the main paper by giving other examples of the frames where was possible to obtain the \textcolor[HTML]{DE3A3A}{reference trajectory}. As can be noticed, suitable video clips include frames where different visual features contrast with the whitish appearance of the snow and are evenly distributed over the whole frame.} 
  \label{fig:suitableclips}
\end{figure*}

\subsection{Data and Annotations}
As stated in the main paper, to determine the quality of \algoname's trajectories, we make use of a dedicated evaluation dataset representing real-world application scenarios. The videos belonging to this dataset appear in the test-set of SkiTB.
SkiTB (``Skiers from the Top to the Bottom'') 
 is a video dataset we collected to implement deep learning-based athlete-tracking methods for different skiing disciplines.
 It contains 300 broadcasting videos (for a total of 392K frames) of 196 professional skiers (alpine skiers, ski jumpers, and freestyle skiers) performing their gestures from the start to the finish. Due to the high spatial extent of skiing courses, each performance is captured across several different cameras put in sequence. Each frame of the videos was manually annotated by our research team with a bounding-box containing the appearance of the athletes' bodies and equipment, following the instructions commonly used for the creation of visual tracking datasets \cite{OTB,VOT2021,VOT2022,fan2019lasot,dunnhofer2021first,ijcv}. All videos are labeled with the skiing sub-discipline performed, the course location, as well as the weather conditions.
 100 of such 300 multi-camera videos feature the performance of alpine skiers,
where the first 60 are set as
training set
and the remaining 40 as test-set, considering an ordering based on the date of the competition.  
This setting respects the real-world condition where the learning-based algorithms are applied to test data acquired at a later time than the training data. 
 Such a training set, which comprises 662 monocular videos and 130934 frames, has been used to fine-tune the STARK \cite{Stark} tracker, while the 439 monocular videos present in the test set have been exploited as candidates for the generation of the reference trajectories as described in Section \ref{sec:experimental} of the main paper.
 Figure \ref{fig:suitableclips} gives additional examples of the situations in which was possible to obtain a reference trajectory, and hence evaluate the error committed by \algoname.
 The virtual graphics present in the video frames have been manually labeled with bounding-boxes in order to avoid the influence of their static visual features in the execution of the camera motion estimation step. All the video frames presented in the main paper and in this supplementary document (Figures \ref{fig:qualex}, \ref{fig:pipeline}, \ref{fig:feetpt}, \ref{fig:hist}, \ref{fig:errorvis}, \ref{fig:suitableclips}) are taken from video clips appearing in the dataset used for validating \algoname.

Other already-published datasets \cite{SkiPose,sporri2016reasearch} were not used for their unsuitability to the task of interest. SkiPose2D \cite{SkiPose} provides just 2D images annotated with sparse pose annotations, thus it is not suitable for dense per-frame tracking as it is required by continuous trajectory visualization in each frame. SkiPosePTZ \cite{sporri2016reasearch} was not employed because of its limited representation of real-world scenarios. Such a dataset was collected by capturing six alpine skiers on a single course composed of just three turning gates. Moreover, poles with markers have been densely placed on the side of the track in order to register the video frames to a geo-spatial reference system. Such marking poles, which are clearly visible in the frames, are not present in normal conditions of alpine skiing training or competition. Thus, in SkiPosePTZ, the frame matching methods for camera motion estimation would have had the opportunity to exploit visual features not present in real conditions, ultimately leading to a wrong assessment of the error committed by \algoname.

\subsection{Performance Measures}
In the following, we give more information about the employed performance measures. The proposed Mean Per Point Trajectory Error (\mppte) is defined, for a video clip $\video$, as follows:
\begin{align}
    \text{MPPTE}\downarrow_{\mathcal{V}}\: = \frac{1}{T} \sum_{t=0}^{T-1} \Big( \frac{1}{|\trajgt_t|} \sum_{i \in \mathcal{I}_t} ||\point_i^{(t)} - q_i^{(t)}||_2^2 \Big),
\end{align}
where $q_i^{(t)} \in \trajgt_t$ and $\mathcal{I}_t = \{ i \} \subseteq \{0, \cdots, t\} : q_i^{(t)} = [ x_i^{(t)}, y_i^{(t)}], 0 \leq x_i^{(t)} < w, 0 \leq y_i^{(t)} < h$. $\mathcal{I}_t$ is the set of the indices of points $q_i^{(t)}$ that remain visible in $\frame_t$.
To obtain a single score that summarizes the performance of \algoname, the average $\text{MPPTE}\downarrow_{\mathcal{V}}$ value across all evaluation video clips is used. 
In simple words, in each frame, $\text{MPPTE}\downarrow_{\mathcal{V}}$ computes the Euclidean distance between the points of the reference trajectory $\trajgt_t$ and the corresponding ones, according to the temporal index of insertion, of the predicted trajectory $\traj_t$. 
Overall, this measure gives information on the spatial distance in pixels occurring between the reference trajectory's points that are visible and the respective but predicted by \algoname. 

The \mppte\ does not take into account the fact that the $\point_i^{(t)}$ could be removed from $\traj_t$ because of Eq. (\ref{eq:extremes}). Due to such a condition, $\traj_t$ and $\trajgt_t$ could have different lengths since points that are discarded are not inserted back later on. Thus, to compare trajectories of different lengths we employed Dynamic Time Warping (\dtw) which measures the alignment of two trajectories in such conditions \cite{dtw}. For a video clip $\video$, the score is defined as 
\begin{align}
    \text{DTW}\downarrow_{\mathcal{V}}\: = \frac{1}{T} \sum_{t=0}^{T-1} dtw(\traj_t, \trajgt_t)
\end{align}
where $dtw(\traj_t, \trajgt_t)$ is the function that computes the Dynamic Time Warping distance in pixels between the predicted and reference trajectories. The overall \dtw\ measure is obtained by averaging across all the evaluation videos.

The \mse\ measure used to compare $\homo$ with the respective reference $\mathbf{H}^{(\text{LOFTR})}_t$, was implemented for a video clip $\video$ as:
\begin{align}
    \text{MSE}\downarrow_{\mathcal{V}}\: = \frac{1}{T} \sum_{t=0}^{T-1} \Big(\frac{1}{9} ||\homo - \mathbf{H}^{(\text{LOFTR})}_t||_2^2 \Big).
\end{align}
As for the \mppte\ and \dtw\ measures, the average across all the video clips was employed to obtain a single score.

For further details about the \auc\ measure, please see \cite{OTB}.

\subsection{Implementation Details}
To obtain the fine-tuned version of STARK \cite{Stark} (STARK-ft), we exploited the frames of the aforementioned training set. We used the original code provided by the authors to adapt the STARK-ST50 model pre-trained for generic object tracking. Except for the number of epochs in the stage-one training (which was set to 200), default values have been used for the  hyper-parameters. Also for the SiamRPN++ \cite{siamrpn} and SuperDiMP \cite{DiMP,VOT2020} visual trackers we used the code provided by the authors along with their pre-trained models. MOSSE \cite{MOSSE} instead was implemented by the pyCFtracking library \cite{pycftrackers}. 
Regarding the camera motion tracking estimation methods, the code provided by the authors was used to implement SuperPoint \cite{SuperPoint} and SuperGlue \cite{SuperGlue} with the pre-trained weights for outdoor conditions. The Kornia library \cite{kornia} was instead used to implement LOFTR \cite{loftr} (the instance for outdoor environments has been exploited). For the ST + LK \cite{shitomasi,lk} and ORB + BF \cite{orb} methods as well as for RANSAC \cite{ransac} the OpenCV implementations \cite{opencv} were employed.

\section{Additional Results}

\input{tables/sharpen.tex}

\paragraph{Sharpening Frames.} Considering the limited availability of visual features in video frames capturing winter scenarios, we explored the impact of the following convolutional filter 
\begin{align}
    S = \begin{bmatrix}
-1 & -1 & -1\\
-1 & 9 & -1 \\
-1 & -1 & -1
\end{bmatrix}
\end{align}
and of the Unsharp mask to sharpen each video frame before the key-point ST + LK detection and tracking step. The results are reported in Table \ref{tab:sharpen} and show that a $3\times3$ filter slightly improves the trajectories in terms of \mppte. The other filter instead reduces the performance.

\paragraph{Initialization with a Skier Detector.}
We performed experiments to understand the impact on \algoname\ of an initialization bounding-box given by a detector rather than a human annotator. In such initialization conditions, \algoname\ becomes a fully automatic system that does not require any human intervention. To implement the detector, we used a YOLOv5x instance \cite{yolov5} fine-tuned (with default hyper-parameters) on the same SkiTB training-set used to fine-tune STARK-ft. With the initialization given by the detector, the \algoname's configuration with STARK-ft as skier tracker and ST + LK  as camera motion estimation method achieves \mppte\ and \dtw\ errors of 15.8 and 344 pixels, respectively. The quality of the initial annotation hence influences negatively the trajectory visualization, since the two measures degrade by 3.7 and 31 pixels, respectively. We hypothesize that this behavior is due to the visual object tracker when initialized with the detector's localization. Such a noisy bounding-box prejudices the initialization step that builds models of the target skier, and in turn, such noisy models affect the localization ability of the tracker. 

\begin{figure*}[t]
\centering
  \includegraphics[width=\linewidth]{images/morequalexcompressed.pdf}
  \caption{\textbf{Aesthetically pleasant trajectory visualizations.} This figure shows examples of enhanced visualizations based on the trajectory generated by \algoname\ for alpine skiers, snowboarders, and ski jumpers. In these cases, in each frame of the video clips, the \textcolor[HTML]{4B8ED1}{trajectories} were smoothed using the Savitzky–Golay filter \cite{savitzky1964smoothing} and the trajectory points falling inside the area defined by the bounding-box $\bbox_t$ were filtered out.} 
  \label{fig:morequalex}
\end{figure*}

\paragraph{More Qualitative Examples.} Figure \ref{fig:morequalex} shows enhanced trajectory visualizations for different disciplines and application scenarios. Additional qualitative videos showing the trajectories generated by \algoname\ can be reached through this link \videolink.

\section{Limitations and Future Work}
We think \algoname\ can serve as a baseline for future research on trajectory visualization and reconstruction in monocular video-based skiing performance analytics. We hypothesize that the capabilities of \algoname\ could be improved by better integrating the different modules of the pipeline, and potentially through an end-to-end optimization stage of the learning modules and backbone networks involved. Datasets to train such an approach should be also investigated.
The system could be also enhanced by exploiting human pose trackers instead of bounding-box ones. 
A human pose representation could be exploited to compute a more consistent point of contact between the athlete and the snow surface. Furthermore, the motion modeling of the different human body parts could enable the development of solutions able to simultaneously reconstruct the trajectory of disparate parts of the athlete (e.g. hands or feet) or of its equipment (e.g. skis). This direction should be investigated in parallel with research for more accurate human pose detection and tracking in skiing.
Finally, we think that the better exploitation of the specific cues appearing on the slope and in training/competition scenarios could lead to an enhanced trajectory visualization. Methods that extend the applicability conditions of the proposed \algoname\ to situations in which the employed LOFTR-based validation process failed, should be also studied in the future. In some contexts (e.g. for broadcasting applications), such a process could also exploit the repetitive movements of camera operators that follow different skiers during competitions on the same track to develop more accurate camera motion tracking models.

%% file: tables/sharpen.tex
\begin{table}[t]

\fontsize{7}{8}\selectfont
	\centering
	\caption{\algoname\textbf{'s performance when sharpening filters are applied.} A small filter of size $3\times3$ slightly improves the \mppte\ results, while the Unsharp mask slightly decreases the quality of trajectories.}
	\label{tab:sharpen}
	\setlength\tabcolsep{.25cm}
    \rowcolors{2}{tblrowcolor2}{tblrowcolor1}
    \begin{tabular}{l | c c c}
		\toprule
        Sharpening & w/o & filter $S$ & Unsharp mask \\
        \midrule
        \mppte  & 12.1 & 11.6 & 12.2 \\
        \dtw & 313 & 314 & 343 \\
	
		\bottomrule		
\end{tabular}

\end{table}

%% file: PaperForReview.bbl
\begin{thebibliography}{10}\itemsep=-1pt

\bibitem{SkiPose}
Roman Bachmann, J{\"o}rg Sp{\"o}rri, Pascal Fua, and Helge Rhodin.
\newblock Motion capture from pan-tilt cameras with unknown orientation.
\newblock In {\em 2019 International Conference on 3D Vision (3DV)}, pages
  308--317. IEEE, 2019.

\bibitem{DiMP}
Goutam Bhat, Martin Danelljan, Luc~Van Gool, and Radu Timofte.
\newblock Learning discriminative model prediction for tracking.
\newblock In {\em Proceedings of the IEEE/CVF international conference on
  computer vision}, pages 6182--6191, 2019.

\bibitem{MOSSE}
David~S Bolme, J~Ross Beveridge, Bruce~A Draper, and Yui~Man Lui.
\newblock Visual object tracking using adaptive correlation filters.
\newblock In {\em 2010 IEEE computer society conference on computer vision and
  pattern recognition}, pages 2544--2550. IEEE, 2010.

\bibitem{opencv}
G. Bradski.
\newblock {The OpenCV Library}.
\newblock {\em Dr. Dobb's Journal of Software Tools}, 2000.

\bibitem{stitching}
Matthew Brown and David~G Lowe.
\newblock Automatic panoramic image stitching using invariant features.
\newblock {\em International journal of computer vision}, 74:59--73, 2007.

\bibitem{trajski}
Congying Cai and Xiaolan Yao.
\newblock Dynamic analysis and trajectory optimization for the nonlinear
  ski-skier system.
\newblock {\em Control Engineering Practice}, 114:104868, 2021.

\bibitem{cai2020trajectory}
Cong-ying Cai and Xiao-lan Yao.
\newblock Trajectory optimization with constraints for alpine skiers based on
  multi-phase nonlinear optimal control.
\newblock {\em Frontiers of Information Technology \& Electronic Engineering},
  21(10):1521--1534, 2020.

\bibitem{Calandre2021}
Jordan Calandre, Renaud Péteri, Laurent Mascarilla, and Benoit Tremblais.
\newblock Table tennis ball kinematic parameters estimation from non-intrusive
  single-view videos.
\newblock In {\em 2021 International Conference on Content-Based Multimedia
  Indexing (CBMI)}, pages 1--6, 2021.

\bibitem{openpose}
Zhe Cao, Tomas Simon, Shih-En Wei, and Yaser Sheikh.
\newblock Realtime multi-person 2d pose estimation using part affinity fields.
\newblock In {\em Proceedings of the IEEE conference on computer vision and
  pattern recognition}, pages 7291--7299, 2017.

\bibitem{Chen2011}
Hua-Tsung Chen, Chien-Li Chou, Wen-Jiin Tsai, and Suh-Yin Lee.
\newblock 3d ball trajectory reconstruction from single-camera sports video for
  free viewpoint virtual replay.
\newblock In {\em 2011 Visual Communications and Image Processing (VCIP)},
  pages 1--4, 2011.

\bibitem{chen2018player}
Liang-Hua Chen, Chih-Wen Su, and Hsiang-An Hsiao.
\newblock Player trajectory reconstruction for tactical analysis.
\newblock {\em Multimedia Tools and Applications}, 77(23):30475--30486, 2018.

\bibitem{nielsenreport}
The~Nielsen Company.
\newblock {Audi FIS Ski World Cup 2021/22 - TV Media Evaluation Event Summary},
  2022.

\bibitem{SuperPoint}
Daniel DeTone, Tomasz Malisiewicz, and Andrew Rabinovich.
\newblock Superpoint: Self-supervised interest point detection and description.
\newblock In {\em Proceedings of the IEEE Conference on Computer Vision and
  Pattern Recognition Workshops}, pages 224--236, 2018.

\bibitem{dunnhofer2021first}
Matteo Dunnhofer, Antonino Furnari, Giovanni~Maria Farinella, and Christian
  Micheloni.
\newblock Is first person vision challenging for object tracking?
\newblock In {\em Proceedings of the IEEE/CVF International Conference on
  Computer Vision Workshops}, pages 2698--2710, 2021.

\bibitem{ijcv}
Matteo Dunnhofer, Antonino Furnari, Giovanni~Maria Farinella, and Christian
  Micheloni.
\newblock Visual object tracking in first person vision.
\newblock {\em International Journal of Computer Vision}, 131(1):259--283,
  2023.

\bibitem{dunnhofer2022cocolot}
Matteo Dunnhofer and Christian~Micheloni Machine.
\newblock Cocolot: Combining complementary trackers in long-term visual
  tracking.
\newblock In {\em 2022 26th International Conference on Pattern Recognition
  (ICPR)}, pages 5132--5139. IEEE, 2022.

\bibitem{Dunnhofer2019}
Matteo Dunnhofer, Niki Martinel, Gian~Luca Foresti, and Christian Micheloni.
\newblock {Visual Tracking by means of Deep Reinforcement Learning and an
  Expert Demonstrator}.
\newblock In {\em Proceedings of The IEEE/CVF International Conference on
  Computer Vision Workshops}, 2019.

\bibitem{Dunnhofer2020accv}
Matteo Dunnhofer, Niki Martinel, and Christian Micheloni.
\newblock {Tracking-by-Trackers with a Distilled and Reinforced Model}.
\newblock In {\em Asian Conference on Computer Vision}, 2020.

\bibitem{dunnhofer2021weakly}
Matteo Dunnhofer, Niki Martinel, and Christian Micheloni.
\newblock Weakly-supervised domain adaptation of deep regression trackers via
  reinforced knowledge distillation.
\newblock {\em IEEE Robotics and Automation Letters}, 6(3):5016--5023, 2021.

\bibitem{fan2019lasot}
Heng Fan, Liting Lin, Fan Yang, Peng Chu, Ge Deng, Sijia Yu, Hexin Bai, Yong
  Xu, Chunyuan Liao, and Haibin Ling.
\newblock Lasot: A high-quality benchmark for large-scale single object
  tracking.
\newblock In {\em Proceedings of the IEEE/CVF conference on computer vision and
  pattern recognition}, pages 5374--5383, 2019.

\bibitem{alphapose}
Hao-Shu Fang, Jiefeng Li, Hongyang Tang, Chao Xu, Haoyi Zhu, Yuliang Xiu,
  Yong-Lu Li, and Cewu Lu.
\newblock Alphapose: Whole-body regional multi-person pose estimation and
  tracking in real-time.
\newblock {\em IEEE Transactions on Pattern Analysis and Machine Intelligence},
  2022.

\bibitem{pycftrackers}
fengyang95.
\newblock {pyCFTrackers}, 2019.

\bibitem{ransac}
Martin~A Fischler and Robert~C Bolles.
\newblock Random sample consensus: a paradigm for model fitting with
  applications to image analysis and automated cartography.
\newblock {\em Communications of the ACM}, 24(6):381--395, 1981.

\bibitem{gilgien2013determination}
Matthias Gilgien, J{\"o}rg Sp{\"o}rri, Julien Chardonnens, Josef Kr{\"o}ll, and
  Erich M{\"u}ller.
\newblock Determination of external forces in alpine skiing using a
  differential global navigation satellite system.
\newblock {\em Sensors}, 13(8):9821--9835, 2013.

\bibitem{multiviewgeometry}
Richard Hartley and Andrew Zisserman.
\newblock {\em Multiple View Geometry in Computer Vision}.
\newblock Cambridge University Press, New York, NY, USA, 2 edition, 2003.

\bibitem{resnet}
Kaiming He, Xiangyu Zhang, Shaoqing Ren, and Jian Sun.
\newblock Deep residual learning for image recognition.
\newblock In {\em Proceedings of the IEEE conference on computer vision and
  pattern recognition}, pages 770--778, 2016.

\bibitem{Honda_2022_CVPR}
Yutaro Honda, Rei Kawakami, Ryota Yoshihashi, Kenta Kato, and Takeshi Naemura.
\newblock Pass receiver prediction in soccer using video and players'
  trajectories.
\newblock In {\em Proceedings of the IEEE/CVF Conference on Computer Vision and
  Pattern Recognition (CVPR) Workshops}, pages 3503--3512, June 2022.

\bibitem{Hu2011}
Min-Chun Hu, Ming-Hsiu Chang, Ja-Ling Wu, and Lin Chi.
\newblock Robust camera calibration and player tracking in broadcast basketball
  video.
\newblock {\em IEEE Transactions on Multimedia}, 13(2):266--279, 2011.

\bibitem{yolov5}
Glenn Jocher, Ayush Chaurasia, Alex Stoken, Jirka Borovec, NanoCode012, Yonghye
  Kwon, Kalen Michael, TaoXie, Jiacong Fang, imyhxy, Lorna, Zeng Yifu, Colin
  Wong, Abhiram V, Diego Montes, Zhiqiang Wang, Cristi Fati, Jebastin Nadar,
  Laughing, UnglvKitDe, Victor Sonck, tkianai, yxNONG, Piotr Skalski, Adam
  Hogan, Dhruv Nair, Max Strobel, and Mrinal Jain.
\newblock {ultralytics/yolov5: v7.0 - YOLOv5 SOTA Realtime Instance
  Segmentation}, Nov. 2022.

\bibitem{kotera2019intra}
Jan Kotera, Denys Rozumnyi, Filip Sroubek, and Jiri Matas.
\newblock Intra-frame object tracking by deblatting.
\newblock In {\em Proceedings of the IEEE/CVF International Conference on
  Computer Vision Workshops}, pages 0--0, 2019.

\bibitem{VOT2022}
Matej Kristan, Ale{\v{s}} Leonardis, Ji{\v{r}}{\'i} Matas, Michael Felsberg,
  Roman Pflugfelder, Joni-Kristian K{\"a}m{\"a}r{\"a}inen, Hyung~Jin Chang,
  Martin Danelljan, Luka~{\v{C}}ehovin Zajc, Alan Luke{\v{z}}i{\v{c}}, Ondrej
  Drbohlav, et~al.
\newblock The tenth visual object tracking vot2022 challenge results.
\newblock In {\em Computer Vision -- ECCV 2022 Workshops}, pages 431--460,
  Cham, 2023. Springer Nature Switzerland.

\bibitem{VOT2020}
Matej Kristan, Ale{\v{s}} Leonardis, Ji{\v{r}}{\'\i} Matas, Michael Felsberg,
  Roman Pflugfelder, Joni-Kristian K{\"a}m{\"a}r{\"a}inen, Martin Danelljan,
  Luka~{\v{C}}ehovin Zajc, Alan Luke{\v{z}}i{\v{c}}, Ondrej Drbohlav, et~al.
\newblock The eighth visual object tracking vot2020 challenge results.
\newblock In {\em Computer Vision--ECCV 2020 Workshops: Glasgow, UK, August
  23--28, 2020, Proceedings, Part V 16}, pages 547--601. Springer, 2020.

\bibitem{VOT2021}
Matej Kristan, Ji\v{r}{\'\i} Matas, Ale\v{s} Leonardis, Michael Felsberg, Roman
  Pflugfelder, Joni-Kristian K\"am\"ar\"ainen, Hyung~Jin Chang, Martin
  Danelljan, Luka Cehovin, Alan Luke\v{z}i\v{c}, Ondrej Drbohlav, Jani
  K\"apyl\"a, Gustav H\"ager, Song Yan, Jinyu Yang, Zhongqun Zhang, and Gustavo
  Fern\'andez.
\newblock The ninth visual object tracking vot2021 challenge results.
\newblock In {\em Proceedings of the IEEE/CVF International Conference on
  Computer Vision (ICCV) Workshops}, pages 2711--2738, October 2021.

\bibitem{alexnet}
Alex Krizhevsky, Ilya Sutskever, and Geoffrey~E Hinton.
\newblock Imagenet classification with deep convolutional neural networks.
\newblock {\em Communications of the ACM}, 60(6):84--90, 2017.

\bibitem{kruger2010application}
Andreas Kr{\"u}ger and J{\"u}rgen Edelmann-Nusser.
\newblock Application of a full body inertial measurement system in alpine
  skiing: A comparison with an optical video based system.
\newblock {\em Journal of applied biomechanics}, 26(4):516--521, 2010.

\bibitem{siamrpn}
Bo Li, Wei Wu, Qiang Wang, Fangyi Zhang, Junliang Xing, and Junjie Yan.
\newblock Siamrpn++: Evolution of siamese visual tracking with very deep
  networks.
\newblock In {\em Proceedings of the IEEE/CVF conference on computer vision and
  pattern recognition}, pages 4282--4291, 2019.

\bibitem{tokenpose}
Yanjie Li, Shoukui Zhang, Zhicheng Wang, Sen Yang, Wankou Yang, Shu-Tao Xia,
  and Erjin Zhou.
\newblock Tokenpose: Learning keypoint tokens for human pose estimation.
\newblock In {\em Proceedings of the IEEE/CVF International conference on
  computer vision}, pages 11313--11322, 2021.

\bibitem{Wei2013}
Wei-Lwun Lu, Jo-Anne Ting, James~J. Little, and Kevin~P. Murphy.
\newblock Learning to track and identify players from broadcast sports videos.
\newblock {\em IEEE Transactions on Pattern Analysis and Machine Intelligence},
  35(7):1704--1716, 2013.

\bibitem{lk}
Bruce~D Lucas and Takeo Kanade.
\newblock An iterative image registration technique with an application to
  stereo vision.
\newblock In {\em IJCAI'81: 7th international joint conference on Artificial
  intelligence}, volume~2, pages 674--679, 1981.

\bibitem{Ludwig_2022_WACV}
Katja Ludwig, Philipp Harzig, and Rainer Lienhart.
\newblock Detecting arbitrary intermediate keypoints for human pose estimation
  with vision transformers.
\newblock In {\em Proceedings of the IEEE/CVF Winter Conference on Applications
  of Computer Vision (WACV) Workshops}, pages 663--671, January 2022.

\bibitem{Ludwig_2023_WACV}
Katja Ludwig, Daniel Kienzle, Julian Lorenz, and Rainer Lienhart.
\newblock Detecting arbitrary keypoints on limbs and skis with sparse partly
  correct segmentation masks.
\newblock In {\em Proceedings of the IEEE/CVF Winter Conference on Applications
  of Computer Vision (WACV) Workshops}, pages 461--470, January 2023.

\bibitem{Maksai_2016_CVPR}
Andrii Maksai, Xinchao Wang, and Pascal Fua.
\newblock What players do with the ball: A physically constrained interaction
  modeling.
\newblock In {\em Proceedings of the IEEE Conference on Computer Vision and
  Pattern Recognition (CVPR)}, June 2016.

\bibitem{Matsumura2021}
Seiji Matsumura, Dan Mikami, Naoki Saijo, and Makio Kashino.
\newblock Spatiotemporal motion synchronization for snowboard big air, 2021.

\bibitem{ostrek2019existing}
Mirela Ostrek, Helge Rhodin, Pascal Fua, Erich M{\"u}ller, and J{\"o}rg
  Sp{\"o}rri.
\newblock Are existing monocular computer vision-based 3d motion capture
  approaches ready for deployment? a methodological study on the example of
  alpine skiing.
\newblock {\em Sensors}, 19(19):4323, 2019.

\bibitem{Perse2009}
Matej Perše, Matej Kristan, Stanislav Kovačič, Goran Vučkovič, and Janez
  Perš.
\newblock A trajectory-based analysis of coordinated team activity in a
  basketball game.
\newblock {\em Computer Vision and Image Understanding}, 113(5):612--621, 2009.
\newblock Computer Vision Based Analysis in Sport Environments.

\bibitem{fasterrcnn}
Shaoqing Ren, Kaiming He, Ross Girshick, and Jian Sun.
\newblock Faster r-cnn: Towards real-time object detection with region proposal
  networks.
\newblock {\em Advances in neural information processing systems}, 28, 2015.

\bibitem{rhodin2018learning}
Helge Rhodin, J{\"o}rg Sp{\"o}rri, Isinsu Katircioglu, Victor Constantin,
  Fr{\'e}d{\'e}ric Meyer, Erich M{\"u}ller, Mathieu Salzmann, and Pascal Fua.
\newblock Learning monocular 3d human pose estimation from multi-view images.
\newblock In {\em Proceedings of the IEEE conference on computer vision and
  pattern recognition}, pages 8437--8446, 2018.

\bibitem{kornia}
E. Riba, D. Mishkin, D. Ponsa, E. Rublee, and G. Bradski.
\newblock Kornia: an open source differentiable computer vision library for
  pytorch.
\newblock In {\em Winter Conference on Applications of Computer Vision}, 2020.

\bibitem{orb}
Ethan Rublee, Vincent Rabaud, Kurt Konolige, and Gary Bradski.
\newblock Orb: An efficient alternative to sift or surf.
\newblock In {\em 2011 International conference on computer vision}, pages
  2564--2571. Ieee, 2011.

\bibitem{dtw}
H. Sakoe and S. Chiba.
\newblock Dynamic programming algorithm optimization for spoken word
  recognition.
\newblock {\em IEEE Transactions on Acoustics, Speech, and Signal Processing},
  26(1):43--49, 1978.

\bibitem{SuperGlue}
Paul-Edouard Sarlin, Daniel DeTone, Tomasz Malisiewicz, and Andrew Rabinovich.
\newblock Superglue: Learning feature matching with graph neural networks.
\newblock In {\em Proceedings of the IEEE/CVF conference on computer vision and
  pattern recognition}, pages 4938--4947, 2020.

\bibitem{savitzky1964smoothing}
Abraham Savitzky and Marcel~JE Golay.
\newblock Smoothing and differentiation of data by simplified least squares
  procedures.
\newblock {\em Analytical chemistry}, 36(8):1627--1639, 1964.

\bibitem{shitomasi}
Jianbo Shi and Tomasi.
\newblock Good features to track.
\newblock In {\em 1994 Proceedings of IEEE Conference on Computer Vision and
  Pattern Recognition}, pages 593--600, 1994.

\bibitem{icrfis}
International Ski and Snowboard~Federation (FIS).
\newblock {The International Ski Competition Rules (ICR) Book IV Joint
  Regulations For Alpine Skiing}, 2022.

\bibitem{sporri2016reasearch}
J{\"o}rg Sp{\"o}rri.
\newblock Reasearch dedicated to sports injury prevention-the’sequence of
  prevention’on the example of alpine ski racing.
\newblock {\em Habilitation with Venia Docendi in Biomechanics}, 1(2):7, 2016.

\bibitem{Skimovie}
Philip Steinkellner and Klaus Schöffmann.
\newblock Evaluation of object detection systems and video tracking in skiing
  videos.
\newblock In {\em 2021 International Conference on Content-Based Multimedia
  Indexing (CBMI)}, pages 1--6, 2021.

\bibitem{loftr}
Jiaming Sun, Zehong Shen, Yuang Wang, Hujun Bao, and Xiaowei Zhou.
\newblock {LoFTR}: Detector-free local feature matching with transformers.
\newblock {\em CVPR}, 2021.

\bibitem{supej2020methodological}
Matej Supej, J{\"o}rg Sp{\"o}rri, and Hans-Christer Holmberg.
\newblock Methodological and practical considerations associated with
  assessment of alpine skiing performance using global navigation satellite
  systems.
\newblock {\em Frontiers in Sports and Active Living}, 1:74, 2020.

\bibitem{raft}
Zachary Teed and Jia Deng.
\newblock Raft: Recurrent all-pairs field transforms for optical flow.
\newblock In {\em Computer Vision--ECCV 2020: 16th European Conference,
  Glasgow, UK, August 23--28, 2020, Proceedings, Part II 16}, pages 402--419.
  Springer, 2020.

\bibitem{SMTreport}
Laurent Vanat.
\newblock {2022 International Report on Snow \& Mountain Tourism}, 2022.

\bibitem{VATS2023}
Kanav Vats, Pascale Walters, Mehrnaz Fani, David~A. Clausi, and John~S. Zelek.
\newblock Player tracking and identification in ice hockey.
\newblock {\em Expert Systems with Applications}, 213:119250, 2023.

\bibitem{Stepec_2022_WACV}
Dejan \v{S}tepec and Danijel Sko\v{c}aj.
\newblock Video-based ski jump style scoring from pose trajectory.
\newblock In {\em Proceedings of the IEEE/CVF Winter Conference on Applications
  of Computer Vision (WACV) Workshops}, pages 682--690, January 2022.

\bibitem{wandt2022elepose}
Bastian Wandt, James~J Little, and Helge Rhodin.
\newblock Elepose: Unsupervised 3d human pose estimation by predicting camera
  elevation and learning normalizing flows on 2d poses.
\newblock In {\em Proceedings of the IEEE/CVF Conference on Computer Vision and
  Pattern Recognition}, pages 6635--6645, 2022.

\bibitem{wandt2021canonpose}
Bastian Wandt, Marco Rudolph, Petrissa Zell, Helge Rhodin, and Bodo Rosenhahn.
\newblock Canonpose: Self-supervised monocular 3d human pose estimation in the
  wild.
\newblock In {\em Proceedings of the IEEE/CVF Conference on Computer Vision and
  Pattern Recognition}, pages 13294--13304, 2021.

\bibitem{wang2019ai}
Jianbo Wang, Kai Qiu, Houwen Peng, Jianlong Fu, and Jianke Zhu.
\newblock Ai coach: Deep human pose estimation and analysis for personalized
  athletic training assistance.
\newblock In {\em Proceedings of the 27th ACM international conference on
  multimedia}, pages 374--382, 2019.

\bibitem{OTB}
Yi Wu, Jongwoo Lim, and Ming-Hsuan Yang.
\newblock Object tracking benchmark.
\newblock {\em IEEE Transactions on Pattern Analysis and Machine Intelligence},
  37(9):1834--1848, 2015.

\bibitem{Stark}
Bin Yan, Houwen Peng, Jianlong Fu, Dong Wang, and Huchuan Lu.
\newblock Learning spatio-temporal transformer for visual tracking.
\newblock In {\em Proceedings of the IEEE/CVF International Conference on
  Computer Vision (ICCV)}, pages 10448--10457, October 2021.

\bibitem{Zhu2022}
Yulin Zhu and Wei~Qi Yan.
\newblock Ski fall detection from digital images using deep learning.
\newblock In {\em Proceedings of the 5th International Conference on Control
  and Computer Vision}, ICCCV '22, page 70–78, New York, NY, USA, 2022.
  Association for Computing Machinery.

\bibitem{Zwolfer2021}
Michael Zwölfer, Dieter Heinrich, Kurt Schindelwig, Bastian Wandt, Helge
  Rhodin, Joerg Spoerri, and Werner Nachbauer.
\newblock Improved 2d keypoint detection in out-of-balance and fall situations
  -- combining input rotations and a kinematic model, 2021.

\end{thebibliography}
